\pdfoutput=1

\documentclass[11pt]{article}

\usepackage[final]{acl}

\usepackage{times}
\usepackage{latexsym}

\usepackage[T1]{fontenc}

\usepackage[utf8]{inputenc}

\usepackage{microtype}
\usepackage{booktabs}
\usepackage{multirow}
\usepackage{amsmath,amssymb,stmaryrd}   

\usepackage{makecell} 

\usepackage{inconsolata}
\usepackage{duckuments}

\usepackage{graphicx}
\usepackage{enumitem}

\usepackage{tabularx}

\usepackage{cprotect} 

\newcolumntype{L}[1]{>{\raggedright\arraybackslash}p{#1}}

\usepackage[most]{tcolorbox}

\newtcolorbox[auto counter,number within=subsection]{myBox}[3][]{
arc=2.5mm, 
lower separated=false,
fonttitle=\bfseries,
fontupper=\footnotesize, 
colbacktitle=white!10,
coltitle=black!20!black,
enhanced,
attach boxed title to top left={xshift=0.3cm,yshift=-2mm},
colframe=black!20!black,
colback=white
}

\title{Masks and Mimicry: Strategic Obfuscation and Impersonation Attacks on Authorship Verification}

\author{Kenneth Alperin$^1$ \hspace{0.2in}
        Rohan Leekha$^1$ \hspace{0.2in} 
        Adaku Uchendu$^1$ \hspace{0.2in} 
        Trang Nguyen$^1$ \\
        \textbf{Srilakshmi Medarametla$^{2}$} \hspace{0.2in}
        \textbf{Carlos Levya Capote$^{3}$} \hspace{0.2in}
        \textbf{Seth Aycock$^4$} \hspace{0.2in}
        \textbf{Charlie Dagli$^1$} \vspace{0.1in} \\
        $^1$ MIT Lincoln Laboratory, MA, USA,  
        $^2$ The University of Virginia, VA, USA \\
        $^3$ University of Puerto Rico-Mayaguez, PR, 
        $^4$ University of Amsterdam, Netherlands \\
        $\textsuperscript{\textdagger}$\small{ \textbf{Correspondence}: \texttt{kenneth.alperin@ll.mit.edu}}
        \vspace{0.1in} \\
        }

\begin{document}
\maketitle
\begin{abstract}
The increasing use of Artificial Intelligence (AI) technologies,
such as Large Language Models (LLMs) has led to nontrivial improvements
in various tasks, including accurate authorship identification of documents. 
However, while LLMs improve such defense techniques, they also simultaneously provide a vehicle 
for malicious actors to launch new attack vectors. 
To combat this security risk, we evaluate the adversarial robustness of authorship models (specifically an authorship verification model) to potent LLM-based attacks. 
These attacks include untargeted methods - \textit{authorship obfuscation} and targeted methods - \textit{authorship impersonation}. 
For both attacks, the objective is to mask or mimic the writing style of an author while preserving the original texts'
semantics, respectively. 
Thus, we perturb an accurate authorship verification model, 
and achieve maximum attack success rates of 92\% and 78\% 
for both obfuscation and impersonation attacks, respectively. 

\end{abstract}

\section{Introduction}
Recent advances in Large Language Models (LLMs) have led to the generation of texts, that are almost 
indistinguishable from human-written texts. 
Consequently, LLMs, while impressive, have exacerbated the problem of influence operations 
within our information ecosystem \cite{chen2023combating,lucas2023fighting}. 
This is because malicious actors can now generate their content at scale with little cost. 
We define \textit{influence operations} as 
any form of attack (typically the spread of propaganda) that pollutes our information space with the ultimate goal of infringing upon a democracy. 
Unsurprisingly, such covert attacks thrive in sensitive events such as 
elections, wars, pandemics, and periods of civil unrest \cite{steinfeld2022disinformation}.

Therefore to combat this obvious security risk, a computational solution is adopted - Authorship Analysis, 
which is an (automatic) approach to finding the author of a document \cite{nguyen2023improving}.  
These Authorship Analysis tasks, include Authorship Attribution, Authorship Verification, Forensic Analysis, Author Profiling, etc. 
\cite{tyo2022state}. 
While all these tasks have specific advantages and uses, we are interested in Authorship Verification (AV) models, which answer the question: \textit{given two texts, can you predict if they are written by the same author or not?} 
Texts written by the same author are known as True Trials, while texts written by different authors are False Trials. 
Using such AV models, one can combat influence operations, by verifying if two randomly selected texts are written by the same author or not. This defense technique has been successfully proposed by several researchers \cite{tyo2022state,stamatatos2016authorship}, and we find that deep learning-based models tend to perform the best. 

\begin{figure}
    \centering
    \includegraphics[width=1\linewidth]{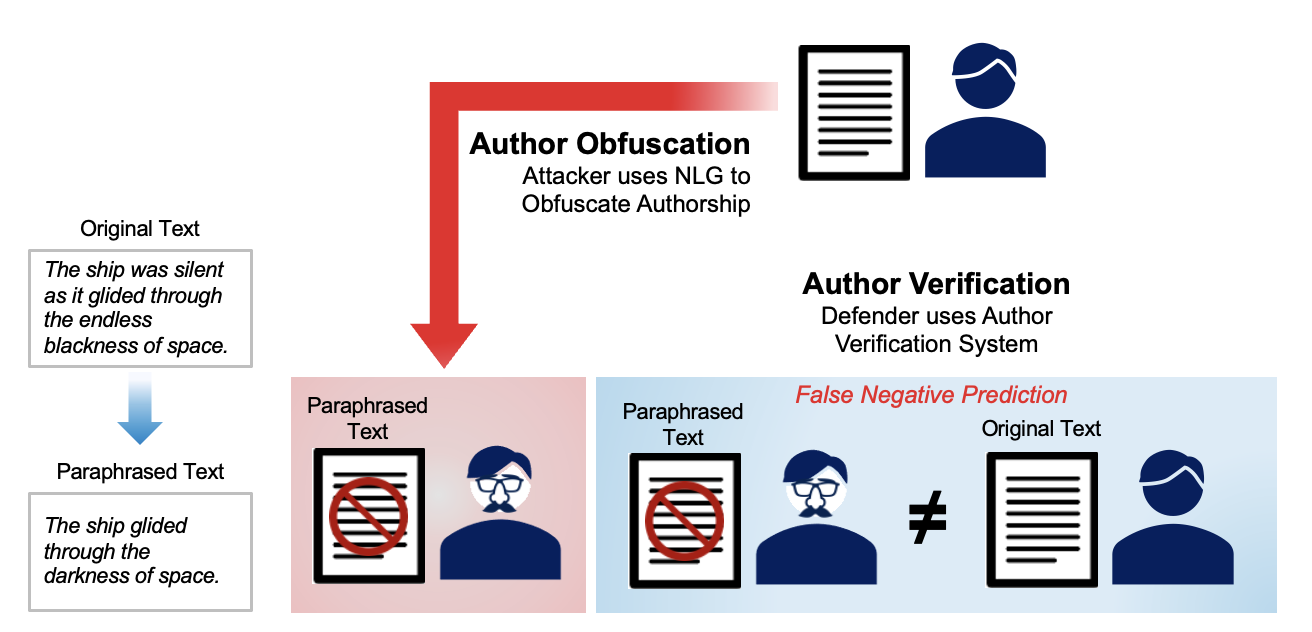}
    \includegraphics[width=1 \linewidth]{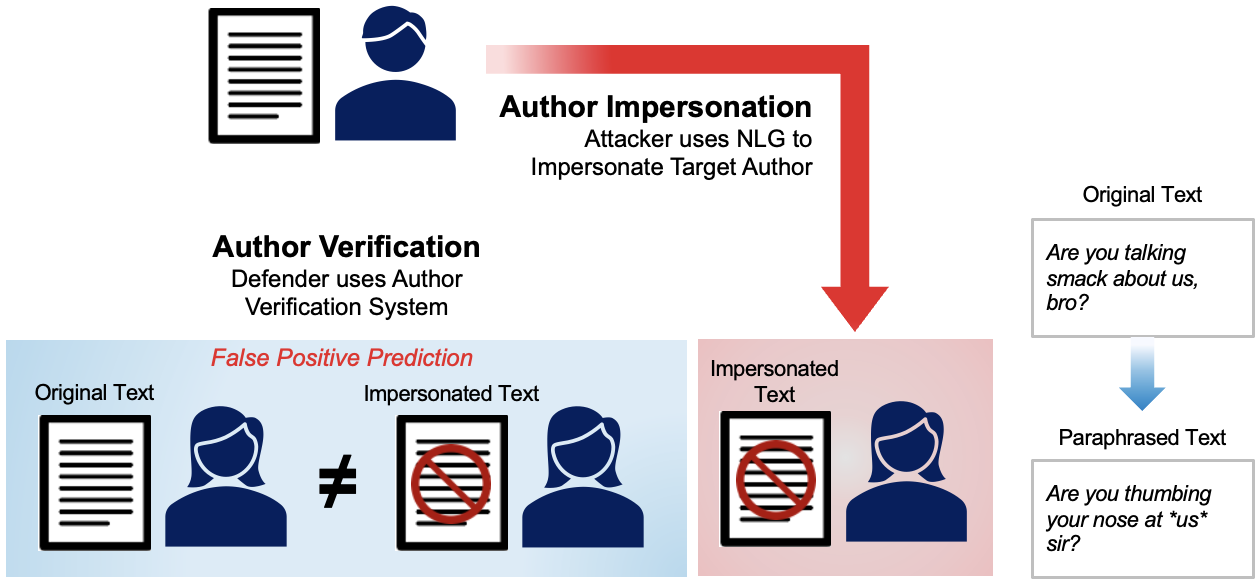}
    \caption{Illustration of \textbf{Authorship Obfuscation} (above) and \textbf{Authorship Impersonation } (below)}
    \label{fig:teaser}
\end{figure}

However, we know that it is not enough to build an accurate AV model, we must evaluate these models under harsher constraints, such as 
realistic adversarial perturbations, specifically \textit{Authorship Obfuscation} and \textit{Authorship Impersonation}. 
\textit{Authorship Obfuscation} is a type of untargeted adversarial attack, with a stronger constraint which is preserving semantics, while 
masking the true authorship of a document \cite{uchendu2023attribution}.  
\textit{Authorship Impersonation}
is a form of targeted adversarial attack, where a target author emulates the writing style of a source author, while 
preserving the semantics of the original texts. 
Adversarial attacks in the context of machine learning are perturbations introduced to 
the model to cause the model to misclassify \cite{goodfellow2014explaining}. 
The goal of these attacks is to perform pre-defined perturbations to achieve mis-classification. 
See Figure \ref{fig:teaser} for illustration of the Authorship Obfuscation and Authorship Impersonation problems. 
Thus, we summarize this study into answering two research questions (RQs): 
\begin{itemize}[noitemsep]
    \item[RQ1:] Can we adversarially perturb an AV model using semantic preserving \underline{untargeted} attacks, known as \textit{Authorship Obfuscation}?
    \item[RQ2:] Can we adversarially perturb an AV model using semantic preserving \underline{targeted} attacks, known as \textit{Authorship Impersonation}?
\end{itemize}

To answer these RQs, we evaluate the adversarial robustness on a 
high-performing AV model - BigBird \cite{nguyen2023improving}, that outperformed strong baselines such as  ELECTRA \cite{clark2020electra}, 
LongFormer \cite{beltagy2020longformer}, and RoBERTa (i.e., DistilRoBERTa) \cite{liu2019roberta}.
Next, we implement several adversarial attacks - obfuscation and impersonation attacks by 
using open-source language models to simulate 
a more realistic scenario of how potential malicious actors will attack AV models in this age of LLMs. 
This yields three language models for the obfuscation attacks - Paraphrasers like 
Mistral\footnote{All Mistral models refer to Mistral-7B-Instruct-v0.1: \url{https://huggingface.co/mistralai/Mistral-7B-Instruct-v0.1}} \cite{jiang2023mistral}, 
DIPPER \cite{krishna2024paraphrasing}, and PEGASUS \cite{zhang2020pegasus};
and three specialized impersonation attack techniques - custom-tuned Mistral, LangChain + RAG\footnote{https://python.langchain.com/docs/tutorials/rag/}, and 
STRAP (GPT-2) \cite{krishna2020reformulating}. 

After probing the AV model with several realistic adversarial 
attacks, we find these attacks 
have a high success rate. 
The obfuscation attacks achieved a maximum attack success rate of 
83\% and 92\% for the two datasets; 
however, for the impersonation we achieved a maximum attack success rate of 78\% when impersonating an author in 
the fanfiction dataset.


\section{Related Work}

\subsection{Authorship Verification (AV)}
Authorship Analysis is an important field for defending against disinformation and malinformation that typically 
aim to mimic the style of a trusted source to increase authenticity.
To combat such security risks, there are two main defense techniques adopted - Authorship Attribution \cite{juola2008authorship,stamatatos2009survey}
and Authorship Verification \cite{stamatatos2016authorship}.
We will focus on Authorship Verification, where researchers have proposed 
stylometric classifiers \cite{seidman2013authorship,weerasinghe2021feature},
statistical-based classifiers \cite{potha2014profile,kocher2017simple,koppel2004authorship,valdez2024team}, 
deep learning-based classifiers \cite{bagnall2015author,nguyen2023improving,singer2023design,tripto2023hansen,boenninghoff2019explainable}, and 
prompt-based techniques \cite{huang2024can,hung2023wrote,ramnath2024cave}.

\subsection{Authorship Obfuscation \& Impersonation }
To assess the robustness of Authorship Analysis models, specifically in the adversarial setting, 
several researchers have proposed author masking techniques, known as Authorship Impersonation and Obfuscation techniques \cite{altakrori2022multifaceted,abegg2023uid,kacmarcik2006obfuscating,brennan2012adversarial,brennan2009practical,le2015secure,emmery2021adversarial,karadzhov2017case,oak2022poster,emmery2024sobr}.
Due to the nontrivial nature of impersonation techniques, most techniques focus on the untargeted author masking approaches (i.e., obfuscation). 
Authorship Impersonation in our context is a variant of the style transfer, where the writing style of a selected author 
is mimicked by another author. 
These techniques include STRAP \cite{krishna2020reformulating}, and others \cite{mir2019evaluating,qi2021mind}. 


More recently, there have been focus on \textit{Paraphrasing attacks} (which involve using language models to rewrite the entire piece of text, while preserving semantics)
as opposed to \textit{classical attacks} \cite{mahmood2019girl,xing2024alison,jin2020bert}
due to the unprecedented benefits of LLMs. 
These paraphrasing attacks include
DIPPER, an encoder-decoder model, T5-XXL that is fine-tuned for paraphrasing \cite{krishna2024paraphrasing}, 
PEGASUS, an encoder-decoder model \cite{zhang2020pegasus}, 
JAMDEC which builds on GPT-2 XL \cite{fisher2024jamdec}, 
and others which use clever prompts to guide desirable generations. 
Researchers have also used LLMs such as 
ChatGPT (GPT-3.5 or GPT-4) \cite{koike2024outfox,macko2024authorship},
BLOOM \cite{bao2024keep}, and more
for paraphrasing documents. 

Finally, \citet{macko2024authorship}, \citet{uchendu2023attribution}, \citet{potthast2016author}, and \citet{altakrori2022evaluation}
survey and comprehensively study the robustness of several obfuscation techniques.

\section{Problem Definitions}

\subsection{Authorship Verification}
AV models aim to answer the question: \textit{given two texts $T_1$ and $T_2$, are they written by the same author or not?}
To verify authorship, if $T_1$ and $T_2$ are written by the same author, we call this a \underline{True Trial}, 
however if they are written by different authors, it is known as a \underline{False Trial}. 

\subsection{Authorship Obfuscation}
In order to evaluate AV models on strong untargeted adversarial perturbations, we adopt several LLMs, as well as several
prompting techniques that make subtle changes to an author's writing style, while 
preserving the semantics. 
Thus, we formally define the obfuscation problem for our context as: 
\begin{myBox}[]{Definition}{AO}
\textsc{\textbf{Definition of Authorship Obfuscation}}.
\textit{Given an Authorship Verification (AV) model $F(x_1, x_2)$
that accurately assigns the label True Trial to 2 pieces of text, $Text_1$ \& $Text_2$ written by the same author,
the AO model $O(x)$ slightly modifies $Text_1$ to $Text^*_1$ (i.e., 
$Text^*_1$ $\leftarrow$ O($Text_1$)) such that the authorship is masked (i.e., 
$F(Text^*_1, Text_2) \neq True~Trial$ or $F(Text^*_1, Text_2) = False~Trial$) and the difference between $Text_1$ and 
$Text^*_1$ is negligible}.
\end{myBox}

This means that a successful obfuscation attack is flipping an accurate prediction of \underline{True Trial} (same author) $\to$ \underline{False Trial} (different authors). 

\subsection{Authorship Impersonation}
To evaluate AV models on strong targeted adversarial perturbations, 
we adopt several customized techniques to transfer style from a source author to a target author, while 
preserving the semantics of the original text. 
We formally, define \textit{Authorship Impersonation} in the context of our task as: 
\begin{myBox}[]{Definition}{authorship impersonation}
\textsc{\textbf{Definition of Authorship Impersonation}}.
\textit{Given an Authorship Verification (AV) model $F(x_1, x_2)$
that accurately assigns the label False Trial to 2 pieces of text,
$Text_1$ to $Text_2$
written by different authors,
the authorship impersonation model, $I(x_{target}, x_{source})$ identifies the target author, $A_{target}$ and source author, $A_{source}$, 
such that $Text^*_{target}$ (i.e., $Text^*_{target}$ $\leftarrow$ I($Text_{target}$, $Text_{source}$)) is written in the same style as $Text_{source}$;
now the authorship is masked (i.e., 
$F(Text^*_{target}, Text_{source}) \neq False~Trial$ or $F(Text^*_{target}, Text_{source} = True~Trial$) and the difference between $Text_{source}$ and 
$Text^*_{source}$ is negligible}.
\end{myBox}

Therefore, a successful attack is defined as flipping an accurate prediction of 
\underline{False Trial} $\to$ \underline{True Trial} as the target author adopts the source author's writing style.


\begin{figure*}
    \centering
    \includegraphics[width=0.7\linewidth]{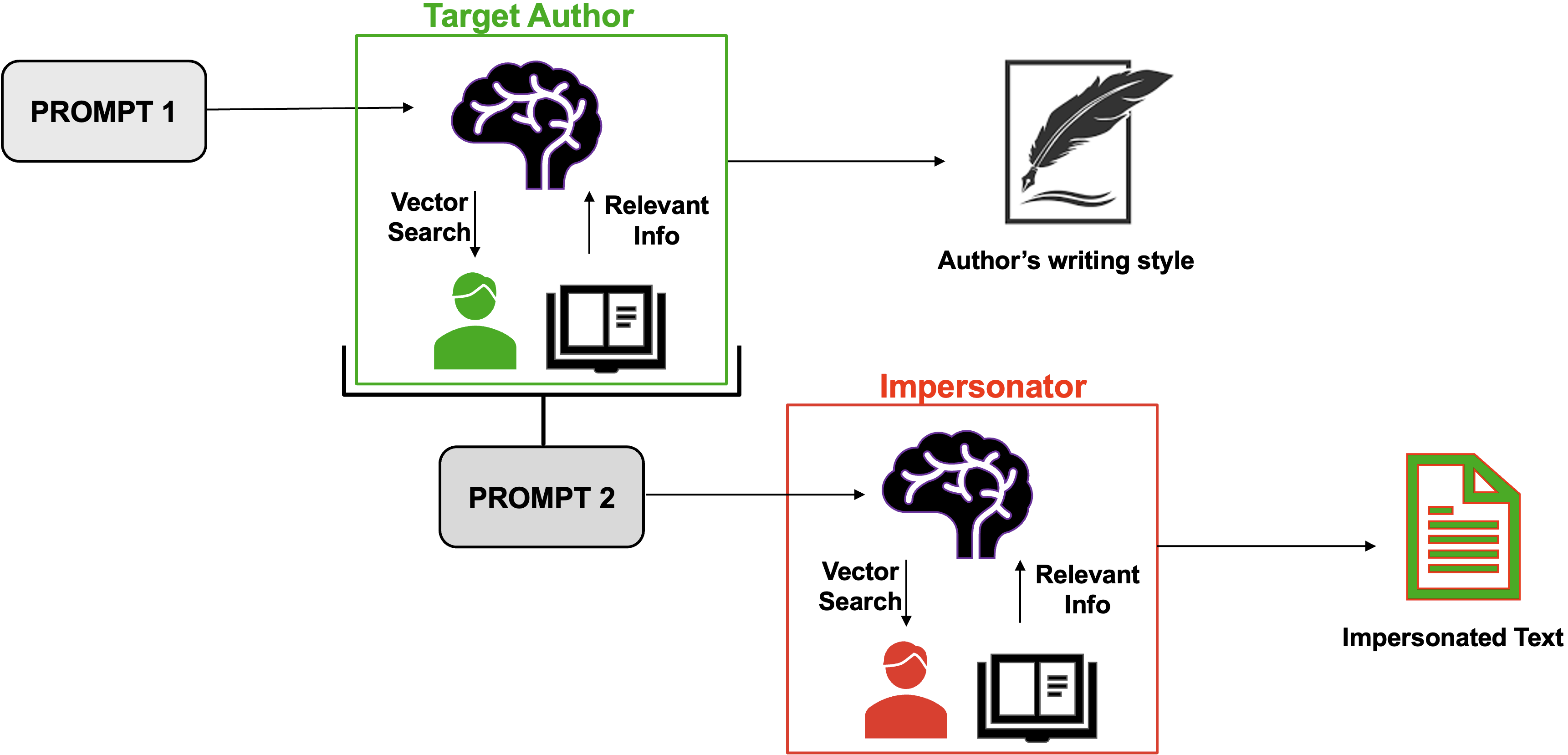}
    \caption{Mistral and RAG framework for Authorship Impersonation. See Figure \ref{fig:rag_mistral} in the Appendix for a more detailed description of the pipeline with prompts}
    \label{fig:aimp_RAG}
\end{figure*}

\section{Methodology}
We evaluate the robustness of \textbf{BigBird} \citet{nguyen2023improving}, a generalizable Authorship Verification (AV) model which outperforms other 
state-of-the-art models, such as ELECTRA \cite{clark2020electra}, 
LongFormer \cite{beltagy2020longformer}, and RoBERTa (i.e., DistilRoBERTa) \cite{liu2019roberta}.
    



\subsection{RQ1: Authorship Obfuscation}
We use the following attacks for obfuscation:
\begin{itemize}[noitemsep]



    \item \textbf{PEGASUS}: is a standard Encoder-Decoder model pre-trained with gap sentences for abstractive summarization \cite{zhang2020pegasus}. 
    However, it is a solid baseline for paraphrasing utilized by several researchers \cite{macko2024authorship}.

    \item \textbf{DIPPER}: is an Encoder-Decoder model - T5-XXL with 11B parameters,  fine-tuned for paraphrasing \cite{krishna2024paraphrasing}.

    \item \textbf{Mistral}: is an instruction-tuned LLM, prompted to paraphrase texts \cite{jiang2023mistral}. 
    See the specific prompts we craft to guide Mistral for obfuscation: 
    \begin{enumerate}[noitemsep]

        \item  \textbf{Vanilla}: Prompting Mistral with the basic instruction to paraphrase the text without using any persona.
        \item \textbf{Zero-shot}: Prompting Mistral to think strategically and paraphrase at most 30\% of the texts.
        \item  \textbf{Step-back}: Prompting Mistral to take a step-back and think strategically. 

        \item \textbf{Author Profile-Aware}: Prompting Mistral to increase the lexical diversity by 60\% by utilizing the following stylistic elements used to write - voice, tone, diction, sentence structure, metaphors \& similes, pacing, imagery, dialogue, age-related features, gender-related features, educational background, psychological traits, cultural \& geographic influences, and social \& occupational factors. 
        

    \end{enumerate}
\end{itemize}
See Table \ref{tab:obf_prompts} in the Appendix for all the prompts we use for the authorship obfuscation attacks.

\subsection{RQ2: Authorship Impersonation}
We perform impersonation attacks, with the following methods: 

\begin{itemize}[noitemsep]
  
    \item \textbf{Mistral and RAG}: We use Retrieval-Augmented Generation (RAG) and the Mistral-7B v0.1 model to perform authorship impersonation by transforming the writing style of a target author into that of the source author. We use a multi-level RAG approach for this pipeline; first we extract and understand the style of the target author, and second apply the style of the target author to rewrite content from the source author without changing the context of the source author. RAG enhances language models by combining retrieval mechanisms with generative capabilities. Instead of relying solely on a model’s internal knowledge, RAG helps in retrieving relevant external information and feeds it into the generation process. 
    This improves accuracy, contextual relevance, and adaptation to specific domains or styles.
    See Figure \ref{fig:aimp_RAG} for an illustration of this impersonation technique. 
    In addition, see Figure \ref{fig:rag_mistral} in Appendix for a more detailed description of Figure \ref{fig:aimp_RAG}. 

    \item \textbf{STRAP}: We perform authorship impersonation using the STRAP (Style Transfer Reformulated as Paraphrasing) framework introduced by \cite{krishna2020reformulating}. The pipeline involves three key phases: paraphrasing with a fine-tuned GPT-2 model, fine-tuning a GPT-2 model on original and paraphrased sentences, and style imputation using the newly fine-tuned GPT-2 model. 
    


    

\end{itemize}

\begin{table*}[!htb]
    \centering
    \resizebox{15cm}{!}{
    \begin{tabular}{|c|c|c|c|c|c|}
    \hline
       \textbf{Dataset}  & \textbf{\# of Authors} & 
       \textbf{\# of Trials} &
       \textbf{\# of Stories} & \textbf{Avg. \# of words} & 
       \textbf{Avg. \# of Sentences} \\
       \hline
        Pan20 FanFiction & 1595 & 13957 & 39588 & 260 & 23 \\
        TwitterCeleb & 129 & 5550 & 5386 & 259 & 20 \\
        \hline
        
    \end{tabular}}
    \caption{Summary statistics of dataset. Both are in English}
    \label{tab:summary_stat}
\end{table*}

\subsection{Evaluation Metrics} \label{Evaluation Metrics}
We evaluate how well our adversarial attacks degrade the performance of the AV model
by utilizing several \textit{performance} and \textit{linguistic} metrics. 
For the performance metrics, we obtain numerical values that represent how well the attack performs and degrades the performance using 
ASR (Attack Success Rate), guided by the 
Equal Error Rate (EER). 
To obtain the EER, we use a DET (Detection Error Trade-off) curve which is a plot of the false rejection rate vs. false acceptance rate to obtain where these rates intersect.
This point of equal errors is known as the EER, and the score at which it occurs was chosen as the threshold for deciding a True and False Trial for our experiments. 
For our task, the EER occurs at a score of $0.29$, so then a score equal or above this operating point is considered a True Trial and below the operating point is a False Trial. Note that the EER value itself is not used. We chose the EER operating point score as our threshold instead of another value, such as $0.5$, so that AV system's errors (false alarms and misses) would be balanced before our attacks. 

Additionally, it is not enough to measure how well the attacks perform on the AV models, we must also 
measure the strength of these attacks. 
This is because an attack could easily modify a piece of text such that it becomes gibberish, thus achieving a high ASR, but losing all relevant meaning. 
Therefore, to measure how well the perturbed texts  preserve the semantics of the original texts, we 
employ linguistic metrics - BLEU, BERTScore, and ROUGE. These metrics measure the semantic consistency between text pairs by comparing the lexical and semantic overlap. 
The scores for all three are between $[0,1]$, such that a
score 
closer to one means high semantic consistency and closer to zero means the text pairs are dissimilar. 

\begin{figure}
    \centering
    \includegraphics[width=0.99\linewidth]{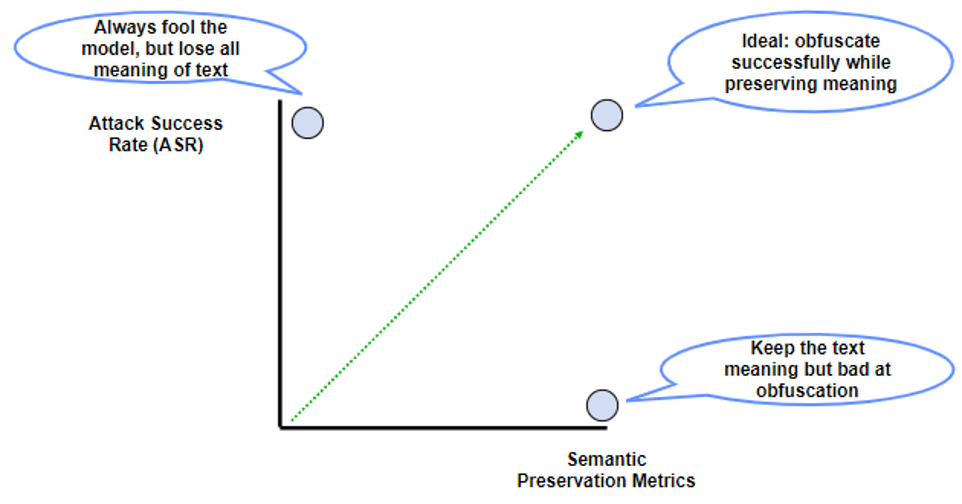}
    \caption{Attack Success Rate (ASR) vs. Semantics}
    \label{fig:asrvs}
\end{figure}

Finally, our objective is to optimize for both high attack success rate and semantic preservation.
See Figure \ref{fig:asrvs} for an illustration of our objective.



\subsection{Dataset Description}
To evaluate the generalizability of the Authorship Obfuscation 
and Impersonation attacks on the BigBird model, we 
compare the performances on two different datasets - 
PAN20 FanFiction and CelebTwitter which are of different domains. For both datasets, we used a closed-set trial design, where the same authors as found in training are also present in test trials but using unseen text data.
Table \ref{tab:summary_stat} contains the summary statistics of the two datasets. 
See description below:
\begin{itemize}
    \item \textbf{PAN20 FanFiction}:
    PAN distributed a dataset for training and testing authorship verification in 2020 \cite{bevendorff20}. This dataset contained True Trial pairs from over 40,000 authors across 1,600 fandoms for training, and from 3,500 authors across 400 fandoms for testing. We downsampled the training data as described by \cite{nguyen2023improving}, and then truncated each author text (originally ~21,000 characters) to approximately 250 tokens, which was identified in \cite{singer2023design} as the minimum sufficient length for evaluation. Truncation was always performed at the end of a sentence, so that no text was cut off.  
    \item \textbf{CelebTwitter}:
    The CelebTwitter trials were created using the PAN 2019 Celebrity Profiling challenge dataset \cite{wiegmann2019celebrity}. The original dataset contained over one million tweets from over 40,000 celebrities. We followed the sampling described in \cite{singer2023design} by first extracting only English tweets from celebrities that also appear in VoxCeleb1 \cite{nagrani2017vox}, and then concatenated a celebrity's tweets together to create a piece of text with a minimum of 250 tokens. 
    
\end{itemize}

\begin{table}[]
    \centering
        \resizebox{7.5cm}{!}{
    \begin{tabular}{|c|c|c|c|c|}
       \toprule 
         \textbf{Obf. Model} &  \textbf{ASR~$\uparrow$} & \textbf{BLEU~$\uparrow$}  & 
         \textbf{ROUGE~$\uparrow$} & \textbf{BERTScore~$\uparrow$}\\
         \midrule
        PEGASUS & 0.06 & 87.6 & 25.4 & 78.6 \\
        DIPPER & 0.80 & 77.1 & 52.6 & 77.0 \\
        \cmidrule{1-5}
        $Mistral_{vanilla}$ & 0.23 & 76.6 & 42.2 & 73.0 \\
        $Mistral_{zeroshot}$ & 0.83 & 70.9 & 44.1 & 76.6 \\
        $Mistral_{stepback}$ & 0.50 & 66.9 & 40.2 & 75.2 \\
        $Mistral_{AP}$ & 0.57 & 67.3 & 36.8 & 69.5 \\
        \bottomrule
        
    \end{tabular}}
    \caption{Authorship Obfuscation Results for Fanfiction}
    \label{tab:ao_results}
\end{table}

\begin{table}
    \centering
        \resizebox{7.5cm}{!}{
    \begin{tabular}{|c|c|c|c|c|}
       \toprule 
         \textbf{Obf. Model} &  \textbf{ASR~$\uparrow$} & \textbf{BLEU~$\uparrow$}  & \textbf{ROUGE~$\uparrow$} &
         \textbf{BERTScore~$\uparrow$}\\
         \midrule
        DIPPER & 0.54 & 75.7 & 49.0 & 75.2 \\
        \hline
        $Mistral_{vanilla}$ & 0.90 & 63.5 & 26.0 & 62.9 \\
        $Mistral_{zeroshot}$ & 0.92 & 67.3 & 20.7 &  63.4 \\ 
        $Mistral_{AP}$ & 0.92 & 71.7 & 29.4 & 65.1 \\
        \bottomrule
        
    \end{tabular}}
    \caption{Authorship Obfuscation Results for CelebTwitter}
    \label{tab:ao_twitter_results}
\end{table}


\section{Authorship Obfuscation Results}

We evaluate the robustness of BigBird \cite{nguyen2023improving} 
to realistic obfuscation attacks in the age of LLMs. 
By using paraphrasers such as DIPPER, PEGASUS, and Mistral, 
we find that DIPPER and Mistral preserve the semantics of the original text, as well as cause the AV model to misclassify at a high rate. 
To evaluate the performance of these attacks, we use the metrics discussed in Section \ref{Evaluation Metrics}. 
Furthermore, in order to investigate the generalizability of our 
attacks, we test the model performance on two datasets of different domains - fanfiction, and celebrity Twitter (now known as X) posts. 

The results for the fanfiction and CelebTwitter datasets are in Tables \ref{tab:ao_results} and \ref{tab:ao_twitter_results}, respectively.
For the fanfiction dataset, we observe that our method - $Mistral_{zeroshot}$ achieved the highest ASR, outperforming the second best obfuscator - DIPPER by 3\%. 
However, DIPPER was able to achieve highest semantic consistency scores on all three metrics, suggesting that it was able to generate obfuscated texts that more closely resemble the original, semantically. 
DIPPER's superior performance, compared to the other baseline - PEGASUS (which underperformed significantly) and Mistral prompts - 
$Mistral_{vanilla}$, $Mistral_{stepback}$, and $Mistral_{AP}$ is 
because DIPPER is the only model that was trained with the objective of paraphrasing, while PEGASUS was trained for abstractive summarization. 
Next, for the other Mistral prompts, only $Mistral_{vanilla}$ achieved a low ASR - 23\%, however, was still able to preserve the semantics decently. The other prompts performed well, achieving nontrivial ASR of 50\% and 57\% for $Mistral_{stepback}$, and $Mistral_{AP}$, respectively.

However, we observe more exaggerated performances by Mistral on the CelebTwitter dataset in 
Table \ref{tab:ao_twitter_results}. 
First, due to the expensive nature of running all the experiments, 
we wanted to compare the top-3 performing attacks - DIPPER, $Mistral_{zeroshot}$, $Mistral_{AP}$. 
Additionally, we include the baseline Mistral prompt - $Mistral_{vanilla}$ to compare the increase in improvements with the other
Mistral prompts. 
We observe that $Mistral_{zeroshot}$ and $Mistral_{AP}$ , achieves the highest ASR - 92\%, outperforming Dipper (54\%) by a large margin, while $Mistral_{vanilla}$ performs comparably, achieving a 90\% ASR. 
This suggests that the performance of the obfuscators could be domain-specific. 
However, as witnessed on the fanfiction dataset, DIPPER consistently outperformed all other models on the semantic metrics.

\begin{table}[]
    \centering
    \resizebox{7.5cm}{!}{
    \begin{tabular}{|c|c|c|c|c|}
    \hline
        \textbf{\makecell{Target \\Author}} & 
        \textbf{\makecell{Target Stories \\for Tuning}} &
        \textbf{\makecell{\# of Source \\Authors}} & 
        \textbf{\makecell{\# of Source \\Stories}} & 
        \textbf{\makecell{False Trial Pairs \\in Test Set}} \\
        \hline
        A & 6 & 208 & 356 & 558 \\
        B & 6 & 217 & 294 & 489 \\
        C & 8 & 185 & 328 & 487 \\
        D & 4 & 201 & 339 & 508 \\
        E & 6 & 221 & 343 & 504 \\
        \hline
    \end{tabular}}
    \caption{Initial Experimental Setup for Authorship Impersonation}
    \label{tab:imp5}
\end{table}


\begin{table}[]
    \centering
    \resizebox{7.5cm}{!}{
    \begin{tabular}{|c|c|c|c|}
    \hline
      \textbf{Target Author}   &  
      \textbf{\# of Stories} &
      \textbf{STRAP-ASR $\uparrow$}  & 
      \textbf{Mistral RAG-ASR $\uparrow$} \\
      \hline
       A  & 6 & 0.50 & \textbf{0.54} \\
       B & 6 & 0.30 & \textbf{0.35} \\
       C & 8 & 0.52 & \textbf{0.75} \\
       D & 4 & 0.11 & \textbf{0.48} \\
       E & 6 & \textbf{0.77} & 0.42 \\
       \hline
    \end{tabular}}
    \caption{Attack Success Rate for Authorship Impersonation}
    \label{tab:ai_5}
\end{table}


\section{Authorship Impersonation Results}

The goal of authorship impersonation is to modify the writing style of an author such that the original author of a document is detected as a particular target author. In obfuscation, we are going from one author to any other author, whereas in impersonation, we are going from any other author to one particular author, making this a much harder problem. 

Table \ref{tab:imp5} shows the details for the initial experimental setup for the impersonation. From the fanfiction dataset, we took the five most prolific authors, and used their stories from the validation set to do the fine-tuning and in-context learning. A key thing to note is in the False Trial pairs for each author in the test set, their stories are compared to stories from hundreds of other authors, so it is a diverse set of documents we are trying to impersonate to a particular author. The defender system is the same BigBird model we used for obfuscation of fanfiction, and our attacker approach is to use the two impersonation techniques, and target stories in the False Trial pairs this time, as we want to fool the model into thinking stories are written by the same author, when in fact they are not. In the False Trial pairs, the same source document can show up in multiple pairs against different target documents, and source authors can have multiple documents in the pairs. We did not showcase impersonation results on the CelebTwitter dataset due to potential data leakage concerns. Since LLMs are trained on large-scale web data, including social media content, they may have already internalized a celebrity’s writing style, making it an unreliable test for our RAG-based approach. Such overlap could inflate performance metrics, undermining the validity of our evaluation.

\begin{figure}[!htb]
    \centering
    \includegraphics[width=1\linewidth]{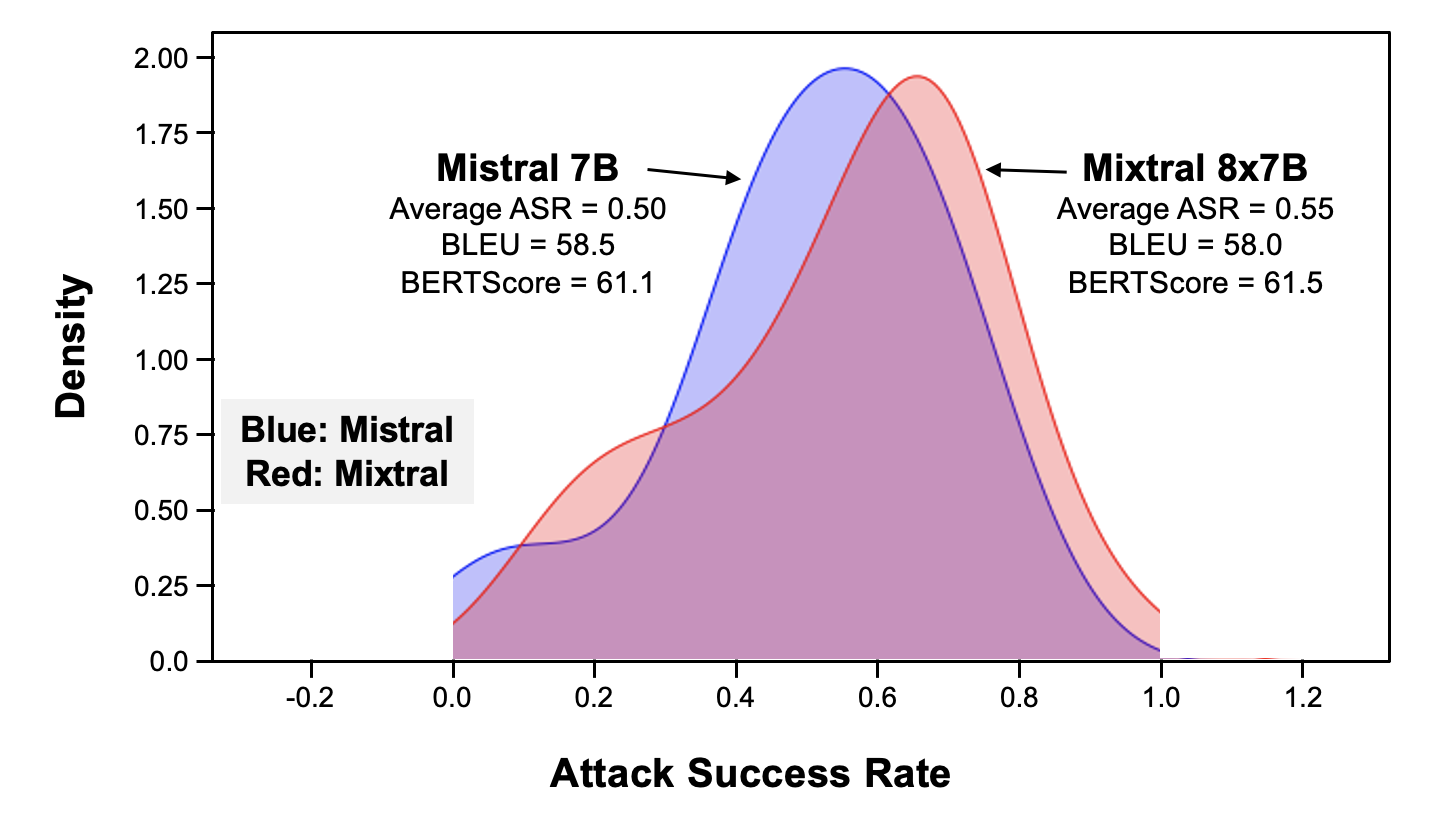}
    \caption{Density Plot of Attack Success Rates for Mistral and Mixtral}
    \label{fig:ai_mis_mix}
\end{figure}

Table \ref{tab:ai_5} shows the ASR of each of the authors for STRAP and Mistral. 
We compare multiple LLMs in the same model family - Mistral-7B v0.1 and Mixtral-8x7B\footnote{\url{https://huggingface.co/mistralai/Mixtral-8x7B-Instruct-v0.1}}. 
Figure \ref{fig:ai_mis_mix} shows the distribution of the ASRs for Mistral and Mixtral. 
The average ASR for Mistral is 50\%, while the average ASR for Mixtral is 55\%. 
This suggests that there is a 55\% chance of taking any other document in the fanfiction and modify it to appear as a document written by the target author, flipping true negatives to false positives. 
The BLEU and BERTScores for both models are within .05 of each other, indicating similar semantic preservation tendencies.

\begin{figure}
    \centering
    \includegraphics[width=1\linewidth]{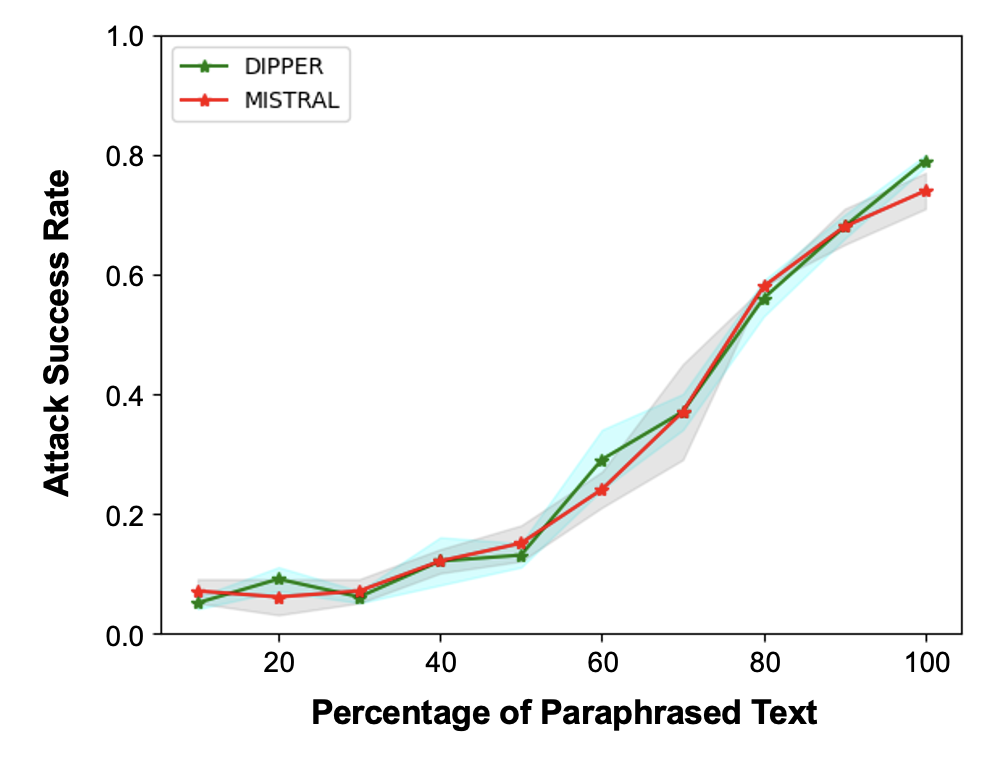}
    \caption{Attack Success Rate (ASR) vs. Percentage of Paraphrased Text}
    \label{fig:ao_abl}
\end{figure}

\section{Ablation Study}

\subsection{Degree of Authorship Obfuscation}
Given that DIPPER and Mistral for some prompts achieved high ASR, 
we wanted to investigate how much of a given text needs to 
be paraphrased to achieve a decent ASR. 
To that end, we conducted an ablation study using DIPPER and $Mistral_{zeroshot}$ to paraphrase a percentage of randomly selected texts of the documents and observe the ASR at those percentages. 
See results shown in Figure \ref{fig:ao_abl}. 

The ablation study is performed 
on the fanfiction dataset for DIPPER and $Mistral_{zeroshot}$.
We observe that the AV models showed robustness to low and medium levels of obfuscation attacks, only starting to degrade in performance 
when about 50-60\% of the document is paraphrased. 
This highlights the robustness of the AV model. 
Thus for future work, it would be interesting to 
not only blindly paraphrase a percentage of texts in a document, 
but to identify the most impactful portions of a document to obfuscate.

\begin{table}[!htb]
    \centering
    \resizebox{7.5cm}{!}{
    \begin{tabular}{|c|c|c|c|}
    \hline
       \textbf{Target Author}  & \textbf{One Story ASR} & 
       \textbf{Three Stories ASR$^*$} & 
       \textbf{All Stories ASR} \\
       \hline
         A & 0.50 & 0.56 & 0.54 \\
         B & 0.37 & 0.65 & 0.35 \\
         C & 0.52 & 0.79 & 0.75 \\
         D & 0.11 & 0.46 & 0.48 \\
         E & 0.78 & 0.59 & 0.42 \\
         \hline
    \end{tabular}}
    \caption{Impersonation Ablation Study *95\% confidence interval of $\pm$ .03}
    \label{tab:ai_abl}
\end{table}


\subsection{In-Context Data for Authorship Impersonation}
Similar to the obfuscation, we ran an ablation study on the initial impersonation experiment to observe the number of stories needed to achieve a successful impersonation attack with the Mistral approach
as shown in Table \ref{tab:ai_abl}. 
We observe that four out of the five authors perform better with three stories than using more stories, 
and for most of them, they achieved the highest ASR with three stories. 
These results suggest that less data is needed for optimal performance for the impersonation attack.

\begin{table*}[]
\footnotesize
    \centering
    \begin{tabular}{|p{0.19\linewidth} | p{0.70\linewidth}|}
    \toprule
      \textbf{Method}   &  \textbf{Obfuscated Texts}\\
      \midrule
      Original &  As Harry Potter bravely confronts the Dark Lord in there ultimate battle at Hogwarts, a blinding burst of light engulfs Voldemort and his Death Eaters, making them vanish.\\
      \hline
       DIPPER  &  Harry Potter confronts the Dark Lord in the final battle at Hogwarts. Blinding light consumes Voldemort and his Death Eaters and they vanish. \\
       PEGASUS & As Harry Potter bravely confronted the Dark Lord in the ultimate battle at Hogwarts, a blinding burst of light made them disappear. \\
       $Mistral\_{vanilla}$ & In the climactic showdown at Hogwarts, Harry fearlessly faces off against the Dark Lord and his minions. Suddenly, a brilliant flash of light descends upon them, causing them to evaporate. \\
       $Mistral\_{zeroshot}$ & Harry Potter bravely faces the Dark Lord in their ultimate Hogwarts fight, causing the enemy's demise.\\
       $Mistral\_{stepback}$ & Harry Potter fearlessly faces the Dark Lord in their dramatic confrontation at Hogwarts, a brilliant flash erupts, engulfing Voldemort and his Death Eaters and causing them to disappear without a trace.\\
       $Mistral\_{AP}$ & With unwavering courage, Harry Potter took on the Dark Lord in their final showdown at Hogwarts. Suddenly, a brilliant flash of light enveloped Voldemort and his Death Eaters, erasing them from existence. \\    
       \bottomrule
    \end{tabular}
    \caption{Examples of Obfuscated Texts}
    \label{tab:obf_examples}
\end{table*}

\begin{table*}[!htb]
\footnotesize
    \centering
    \begin{tabular}{|p{0.15\linewidth} | p{0.70\linewidth}|}
    \toprule
      \textbf{Method}   &  \textbf{Impersonated Texts}\\
      \midrule
      Original  & He sighed with relief. Those papers weren’t important. While he started to get up to see what had happened, he was knocked down by Chix’s wings. \\
      \hline
       STRAP  &  She’d been expecting a little more, but she’d been expecting a little more than that. The papers weren’t even that important. Chix was knocked to the ground when he turned to look what had happened to him. \\
       Mistral + RAG & "Whew! Finally got some time to breathe." Nah, those papers were just fine. Holy crap, Chix's wings were so big and powerful, they almost sent me flying off the damn thing! \\
       
       \bottomrule
    \end{tabular}
    \caption{Examples of Impersonated Texts}
    \label{tab:imp_examples}
\end{table*}

\section{Discussion}

Our results highlights the vulnerabilities of AV models when 
subjected to adversarial attacks through authorship obfuscation and authorship impersonation techniques using LLMs. 
See examples of perturbed texts using these methods in Tables \ref{tab:obf_examples} and \ref{tab:imp_examples} for 
the Obfuscation and Impersonation techniques, respectively. 
While prior studies have demonstrated the effectiveness of deep learning-based AV models \cite{bagnall2015author,nguyen2023improving,singer2023design} in distinguishing between different authors, our results reveal significant weaknesses in these models when they encounter realistic, semantic, and context-preserving adversarial perturbations.

Traditional methods like homoglyph substitution \cite{gao2018black}, backtranslation \cite{keswani2016author,shetty2018a4nt, wang2024defending}, 
and synonym swapping \cite{ren2019generating} 
have shown some success in misleading AV models. 
However, these approaches often fail to maintain the semantic integrity of the text \cite{uchendu2023attribution}, 
which can often generate unnatural or nonsensical outputs, reducing the practical applicability of such attacks. 
In contrast, our zero-shot prompting strategies with Mistral and paraphrasing-based obfuscation techniques such as DIPPER demonstrate that AV models struggle to maintain reliable authorship identification when obfuscated,
even when the meaning and coherence of the text are preserved. Specifically, our results show that Mistral-based obfuscation achieves high ASR while maintaining textual coherence, effectively misleading AV models without compromising the quality or readability of the text, noticeably.
Furthermore, we also observe from Tables \ref{tab:ao_results} and 
\ref{tab:ao_twitter_results}, that the strength of the obfuscation 
technique can be domain-specific. This is because of the writing style difference between the two datasets, where FanFiction uses story writing style and 
the Celeb Twitter dataset uses social media post writing style.

In parallel, the authorship impersonation task, which represents a more challenging targeted attack, seeks to manipulate text to mimic the style of a target author while preserving the original semantics. 
Our RAG pipeline successfully transfers stylistic elements from a target author to source author. 
This multi-step RAG process first retrieves the stylistic properties from a target author's previous writings through chain-of-thought prompting to refine these stylistic transformations while maintaining the original meaning. 
Our evaluation demonstrates that LLM-driven impersonation can deceive even the most robust AV models, achieving high success rates in flipping False Trials to True Trials. 
This effectively makes a target author's writing indistinguishable from that of a source author. Such vulnerabilities raise serious security concerns in areas like academic authorship, forensic linguistics, and online misinformation detection.

\begin{figure}[!htb]
    \centering
    \includegraphics[width=1\linewidth]{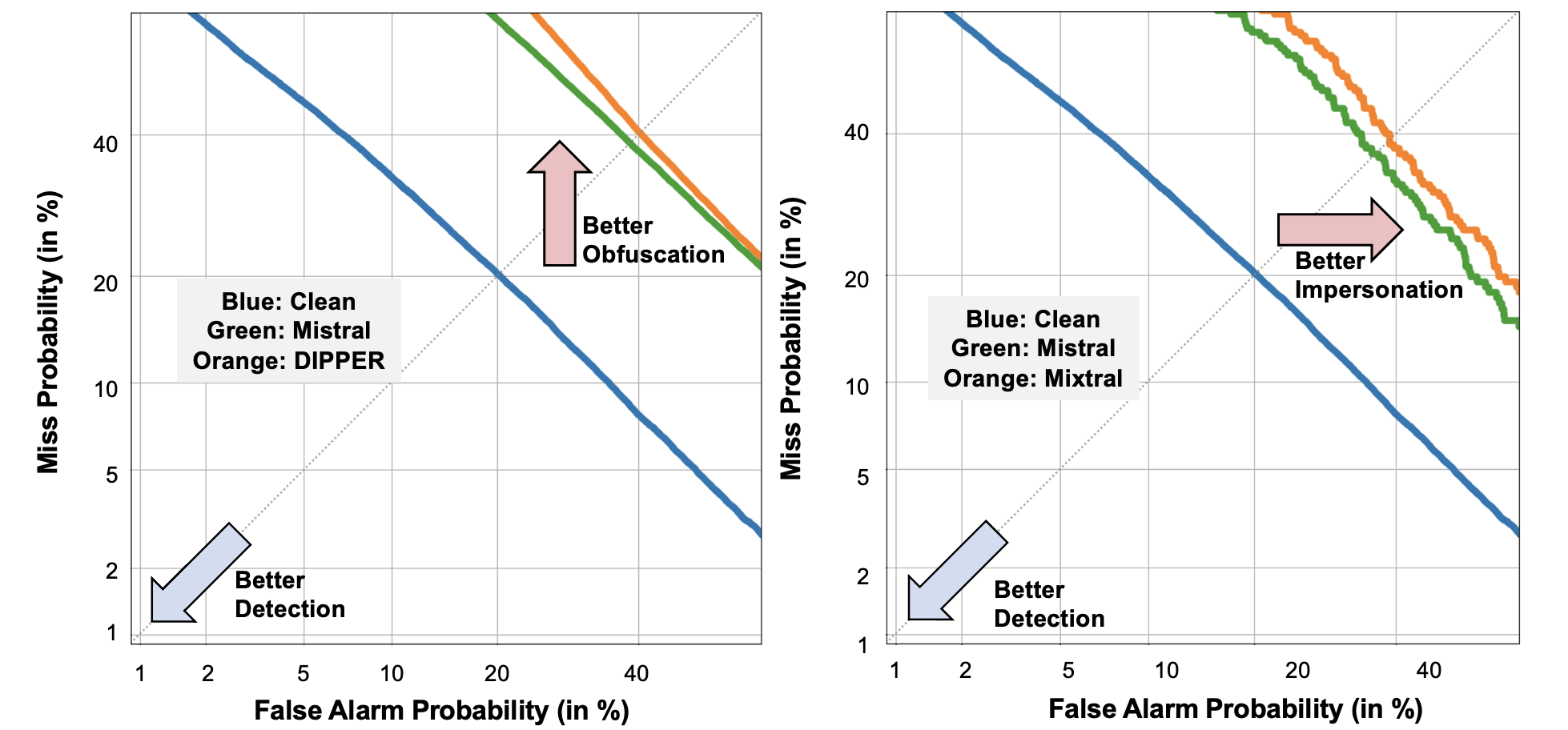}
    \caption{Detection Error Tradeoff Curves for Obfuscation (left) and Impersonation (right) on the fanfiction dataset}
    \label{fig:det}
\end{figure}

Lastly, we observe the strength of the obfuscation and impersonation attacks on the fanfiction datasets by 
plotting a DET Curve' using the misclassification rate as a function of False Alarm Rate 
on a normal log scale. See Figure \ref{fig:det}.
Each line summarizes the performance of a system across a range of thresholds for a given test set. The closer these lines are to the lower-left corner, the better the system's performance.
Thus, we observe that DIPPER \& $Mistral\_{zeroshot}$ for obfuscation and 
Mistral \& Mixtral for impersonation are able to push the verification scores farther away from the unperturbed 
texts scores, such that the AV model misclassifies at a high rate, achieving a successful attack.

\section{Conclusion}
We evaluated the adversarial robustness of a high-performing AV model - BigBird \cite{nguyen2023improving} on adversarial attacks such as 
obfuscation (untargeted) and impersonation (targeted) attacks. 
For the obfuscation attack, we perturbed the first pair of 
accurately predicted True Trials (i.e., same-author documents). 
Therefore, a successful obfuscation attack, flips a True Trial label to False Trial. We achieved high attack success rates with DIPPER, a paraphraser LLM and various prompts used to guide Mistral. 
Next, for the impersonation attack, we aim to perturb the first pair of accurately predicted False Trials (i.e., not-the-same author), 
such that the label is flipped to True Trial. 
Finally, our results expose an alarming security risk which author identification models such as AV models have. 
We are especially alarmed by the results of the impersonation attacks, as these are realistic scenarios in which malicious actors can use to launch devastating security attacks.

\section{Future Work}
In the future, we would expand this work to multilingual datasets, 
to investigate how well our attack techniques can capture an author's style in multiple languages, 
as well as in different domains, such as source code attribution. 
Second, we aim to test out additional LLMs, such as GPT-4 and Claude, to compare newer foundational models for these attacks, and compare our approach to a few-shot learning approach.
Additionally, we aim to fuse more linguistic-based features, 
such as n-gram distribution, with the LLM-based techniques. 
Fourth, we believe implementing an agent-based approach for impersonation could significantly improve the ASR, such as Parameter-Efficient Fine-Tuning (PEFT). 
Finally, to improve our techniques further, we aim to incorporate additional
evaluation metrics, such as machine-generated text detection,
and the goal will be for the generated texts to pass the Turing Test. 
Therefore, our attack method can be used to evaluate the robustness 
of authorship identification models, including authorship verification and attribution, as well as AI-generated text detectors. 

\section{Limitations}
Our research focuses on the usage of LLMs in fooling authorship verification models using non-targeted (Obfuscation) and targeted (Impersonation) approaches on short-and medium form documents and the results may not be universally applicable to 
long form data. 
Additionally, our methodology may face limitations when dealing with multilingual data, which could potentially impact the assessment
of measuring impersonation or obfuscation in these types of datasets. 
Lastly, we do not use LORA or PEFT fine-tuning in either of our methods imitating in accurately
assessing the extend of our attack (impersonation / obfuscation) methods.

\section{Ethical Statement}
Since the advent of LLMs, it is no secret that its abilities are unprecedented for both positive and negative reasons.
Thus, we aim to find the negative ways in which LLMs can be leveraged in the context of authorship identification. 
A famous saying goes - \textit{with great power, comes great responsibility}. 
This means that as we have the knowledge and access to technology that can be used for great good, and great evil, 
it is therefore our 
responsibility to utilize it for great good or at least not cause harm. 
Therefore, while it may seem that we have proposed new attack paradigms, our aim is not 
for malicious use but to create awareness that building an accurate authorship identifier is not enough; 
it must be evaluated under strict constraints such as adversarial perturbations to make sure malicious actors 
are not evading detection. 
Moreover, we achieve successful attacks in a realistic setting using open-source smaller LLMs ($\leq 11$B) which suggests 
that anyone with means can recreate such attacks, at little cost. 
Therefore, we believe that we have fulfilled our responsibility and showcased realistic attack scenarios
that malicious actors may already be using to evade detection. 
Finally, due to the obvious security risk negative applications of LLMs pose, we believe that benefits of this work, outweighs
the risks.

\section*{Acknowledgments}

DISTRIBUTION STATEMENT A. Approved for public release. Distribution is unlimited.
This material is based upon work supported by the Department of the Air Force under Air Force Contract No. FA8702-15-D-0001. Any opinions, findings, conclusions or recommendations expressed in this material are those of the author(s) and do not necessarily reflect the views of the Department of the Air Force.
© 2024 Massachusetts Institute of Technology.
Delivered to the U.S. Government with Unlimited Rights, as defined in DFARS Part 252.227-7013 or 7014 (Feb 2014). Notwithstanding any copyright notice, U.S. Government rights in this work are defined by DFARS 252.227-7013 or DFARS 252.227-7014 as detailed above. Use of this work other than as specifically authorized by the U.S. Government may violate any copyrights that exist in this work.

\bibliography{custom}

\begin{thebibliography}{59}
\providecommand{\natexlab}[1]{#1}

\bibitem[{Abegg(2023)}]{abegg2023uid}
Nicholas Abegg. 2023.
\newblock Uid as a guiding metric for automated authorship obfuscation.
\newblock \emph{arXiv preprint arXiv:2312.03709}.

\bibitem[{Altakrori et~al.(2022)Altakrori, Scialom, Fung, and Cheung}]{altakrori2022multifaceted}
Malik Altakrori, Thomas Scialom, Benjamin~CM Fung, and Jackie Chi~Kit Cheung. 2022.
\newblock A multifaceted framework to evaluate evasion, content preservation, and misattribution in authorship obfuscation techniques.
\newblock In \emph{Proceedings of the 2022 Conference on Empirical Methods in Natural Language Processing}, pages 2391--2406.

\bibitem[{Altakrori(2022)}]{altakrori2022evaluation}
Malik~Hashem Altakrori. 2022.
\newblock \emph{Evaluation Techniques for Authorship Attribution and Obfuscation}.
\newblock Ph.D. thesis, McGill University (Canada).

\bibitem[{Bagnall(2015)}]{bagnall2015author}
Douglas Bagnall. 2015.
\newblock Author identification using multi-headed recurrent neural networks.
\newblock \emph{arXiv preprint arXiv:1506.04891}.

\bibitem[{Bao and Carpuat(2024)}]{bao2024keep}
Calvin Bao and Marine Carpuat. 2024.
\newblock Keep it private: Unsupervised privatization of online text.
\newblock In \emph{Proceedings of the 2024 Conference of the North American Chapter of the Association for Computational Linguistics: Human Language Technologies (Volume 1: Long Papers)}, pages 8670--8685.

\bibitem[{Beltagy et~al.(2020)Beltagy, Peters, and Cohan}]{beltagy2020longformer}
Iz~Beltagy, Matthew~E Peters, and Arman Cohan. 2020.
\newblock Longformer: The long-document transformer.
\newblock \emph{arXiv preprint arXiv:2004.05150}.

\bibitem[{Bevendorff and et~al.(2021)}]{bevendorff20}
Janek Bevendorff and et~al. 2021.
\newblock \href {https://link.springer.com/chapter/10.1007/978-3-030-72240-1_66} {Overview of pan 2021: Authorship verification, profiling hate speech spreaders on twitter, and style change detection}.
\newblock In \emph{Advances in Information Retrieval}, pages 567--573.

\bibitem[{Boenninghoff et~al.(2019)Boenninghoff, Hessler, Kolossa, and Nickel}]{boenninghoff2019explainable}
Benedikt Boenninghoff, Steffen Hessler, Dorothea Kolossa, and Robert~M Nickel. 2019.
\newblock Explainable authorship verification in social media via attention-based similarity learning.
\newblock In \emph{2019 IEEE International Conference on Big Data (Big Data)}, pages 36--45. IEEE.

\bibitem[{Brennan et~al.(2012)Brennan, Afroz, and Greenstadt}]{brennan2012adversarial}
Michael Brennan, Sadia Afroz, and Rachel Greenstadt. 2012.
\newblock Adversarial stylometry: Circumventing authorship recognition to preserve privacy and anonymity.
\newblock \emph{ACM Transactions on Information and System Security (TISSEC)}, 15(3):1--22.

\bibitem[{Brennan and Greenstadt(2009)}]{brennan2009practical}
Michael Brennan and Rachel Greenstadt. 2009.
\newblock Practical attacks against authorship recognition techniques.
\newblock In \emph{21st Innovative Applications of Artificial Intelligence Conference, IAAI-09}, pages 60--65.

\bibitem[{Chen and Shu(2024)}]{chen2023combating}
Canyu Chen and Kai Shu. 2024.
\newblock Combating misinformation in the age of llms: Opportunities and challenges.
\newblock \emph{AI Magazine}, 45(3):354--368.

\bibitem[{Clark et~al.(2020)Clark, Luong, Le, and Manning}]{clark2020electra}
Kevin Clark, Minh-Thang Luong, Quoc~V. Le, and Christopher~D. Manning. 2020.
\newblock \href {https://openreview.net/pdf?id=r1xMH1BtvB} {{ELECTRA}: Pre-training text encoders as discriminators rather than generators}.
\newblock In \emph{ICLR}.

\bibitem[{Emmery et~al.(2021)Emmery, K{\'a}d{\'a}r, and Chrupa{\l}a}]{emmery2021adversarial}
Chris Emmery, {\'A}kos K{\'a}d{\'a}r, and Grzegorz Chrupa{\l}a. 2021.
\newblock Adversarial stylometry in the wild: Transferable lexical substitution attacks on author profiling.
\newblock In \emph{Proceedings of the 16th Conference of the European Chapter of the Association for Computational Linguistics: Main Volume}, pages 2388--2402.

\bibitem[{Emmery et~al.(2024)Emmery, Miotto, Kramp, and Kleinberg}]{emmery2024sobr}
Chris Emmery, Maril{\`u} Miotto, Sergey Kramp, and Bennett Kleinberg. 2024.
\newblock Sobr: A corpus for stylometry, obfuscation, and bias on reddit.
\newblock In \emph{Proceedings of the 2024 Joint International Conference on Computational Linguistics, Language Resources and Evaluation (LREC-COLING 2024)}, pages 14967--14983.

\bibitem[{Fisher et~al.(2024)Fisher, Lu, Jung, Jiang, Harchaoui, and Choi}]{fisher2024jamdec}
Jillian Fisher, Ximing Lu, Jaehun Jung, Liwei Jiang, Zaid Harchaoui, and Yejin Choi. 2024.
\newblock Jamdec: Unsupervised authorship obfuscation using constrained decoding over small language models.
\newblock \emph{arXiv preprint arXiv:2402.08761}.

\bibitem[{Gao et~al.(2018)Gao, Lanchantin, Soffa, and Qi}]{gao2018black}
Ji~Gao, Jack Lanchantin, Mary~Lou Soffa, and Yanjun Qi. 2018.
\newblock Black-box generation of adversarial text sequences to evade deep learning classifiers.
\newblock In \emph{2018 IEEE Security and Privacy Workshops (SPW)}, pages 50--56. IEEE.

\bibitem[{Goodfellow et~al.(2014)Goodfellow, Shlens, and Szegedy}]{goodfellow2014explaining}
Ian~J Goodfellow, Jonathon Shlens, and Christian Szegedy. 2014.
\newblock Explaining and harnessing adversarial examples.
\newblock \emph{arXiv preprint arXiv:1412.6572}.

\bibitem[{Huang et~al.(2024)Huang, Chen, and Shu}]{huang2024can}
Baixiang Huang, Canyu Chen, and Kai Shu. 2024.
\newblock Can large language models identify authorship?
\newblock In \emph{Findings of the Association for Computational Linguistics: EMNLP 2024}, pages 445--460.

\bibitem[{Hung et~al.(2023)Hung, Hu, Hu, and Lee}]{hung2023wrote}
Chia-Yu Hung, Zhiqiang Hu, Yujia Hu, and Roy Lee. 2023.
\newblock Who wrote it and why? prompting large-language models for authorship verification.
\newblock In \emph{Findings of the Association for Computational Linguistics: EMNLP 2023}, pages 14078--14084.

\bibitem[{Jiang et~al.(2023)Jiang, Sablayrolles, Mensch, Bamford, Chaplot, Casas, Bressand, Lengyel, Lample, Saulnier et~al.}]{jiang2023mistral}
Albert~Q Jiang, Alexandre Sablayrolles, Arthur Mensch, Chris Bamford, Devendra~Singh Chaplot, Diego de~las Casas, Florian Bressand, Gianna Lengyel, Guillaume Lample, Lucile Saulnier, et~al. 2023.
\newblock Mistral 7b.
\newblock \emph{arXiv preprint arXiv:2310.06825}.

\bibitem[{Jin et~al.(2020)Jin, Jin, Zhou, and Szolovits}]{jin2020bert}
Di~Jin, Zhijing Jin, Joey~Tianyi Zhou, and Peter Szolovits. 2020.
\newblock Is bert really robust? a strong baseline for natural language attack on text classification and entailment.
\newblock In \emph{Proceedings of the AAAI conference on artificial intelligence}, volume~34, pages 8018--8025.

\bibitem[{Juola et~al.(2008)}]{juola2008authorship}
Patrick Juola et~al. 2008.
\newblock Authorship attribution.
\newblock \emph{Foundations and Trends{\textregistered} in Information Retrieval}, 1(3):233--334.

\bibitem[{Kacmarcik and Gamon(2006)}]{kacmarcik2006obfuscating}
Gary Kacmarcik and Michael Gamon. 2006.
\newblock Obfuscating document stylometry to preserve author anonymity.
\newblock In \emph{Proceedings of the COLING/ACL 2006 Main Conference Poster Sessions}, pages 444--451.

\bibitem[{Karadzhov et~al.(2017)Karadzhov, Mihaylova, Kiprov, Georgiev, Koychev, and Nakov}]{karadzhov2017case}
Georgi Karadzhov, Tsvetomila Mihaylova, Yasen Kiprov, Georgi Georgiev, Ivan Koychev, and Preslav Nakov. 2017.
\newblock The case for being average: A mediocrity approach to style masking and author obfuscation: (best of the labs track at clef-2017).
\newblock In \emph{Experimental IR Meets Multilinguality, Multimodality, and Interaction: 8th International Conference of the CLEF Association, CLEF 2017, Dublin, Ireland, September 11--14, 2017, Proceedings 8}, pages 173--185. Springer.

\bibitem[{Keswani et~al.(2016)Keswani, Trivedi, Mehta, and Majumder}]{keswani2016author}
Yashwant Keswani, Harsh Trivedi, Parth Mehta, and Prasenjit Majumder. 2016.
\newblock Author masking through translation.
\newblock \emph{CLEF (Working Notes)}, 1609:890--894.

\bibitem[{Kocher and Savoy(2017)}]{kocher2017simple}
Mirco Kocher and Jacques Savoy. 2017.
\newblock A simple and efficient algorithm for authorship verification.
\newblock \emph{Journal of the Association for Information Science and Technology}, 68(1):259--269.

\bibitem[{Koike et~al.(2024)Koike, Kaneko, and Okazaki}]{koike2024outfox}
Ryuto Koike, Masahiro Kaneko, and Naoaki Okazaki. 2024.
\newblock Outfox: Llm-generated essay detection through in-context learning with adversarially generated examples.
\newblock In \emph{Proceedings of the AAAI Conference on Artificial Intelligence}, volume~38, pages 21258--21266.

\bibitem[{Koppel and Schler(2004)}]{koppel2004authorship}
Moshe Koppel and Jonathan Schler. 2004.
\newblock Authorship verification as a one-class classification problem.
\newblock In \emph{Proceedings of the twenty-first international conference on Machine learning}, page~62.

\bibitem[{Krishna et~al.(2024)Krishna, Song, Karpinska, Wieting, and Iyyer}]{krishna2024paraphrasing}
Kalpesh Krishna, Yixiao Song, Marzena Karpinska, John Wieting, and Mohit Iyyer. 2024.
\newblock Paraphrasing evades detectors of ai-generated text, but retrieval is an effective defense.
\newblock \emph{Advances in Neural Information Processing Systems}, 36.

\bibitem[{Krishna et~al.(2020)Krishna, Wieting, and Iyyer}]{krishna2020reformulating}
Kalpesh Krishna, John Wieting, and Mohit Iyyer. 2020.
\newblock Reformulating unsupervised style transfer as paraphrase generation.
\newblock In \emph{Proceedings of the 2020 Conference on Empirical Methods in Natural Language Processing (EMNLP)}, pages 737--762.

\bibitem[{Le et~al.(2015)Le, Safavi-Naini, and Galib}]{le2015secure}
Hoi Le, Reihaneh Safavi-Naini, and Asadullah Galib. 2015.
\newblock Secure obfuscation of authoring style.
\newblock In \emph{Information Security Theory and Practice: 9th IFIP WG 11.2 International Conference, WISTP 2015, Heraklion, Crete, Greece, August 24-25, 2015. Proceedings 9}, pages 88--103. Springer.

\bibitem[{Liu et~al.(2019)Liu, Ott, Goyal, Du, Joshi, Chen, Levy, Lewis, Zettlemoyer, and Stoyanov}]{liu2019roberta}
Yinhan Liu, Myle Ott, Naman Goyal, Jingfei Du, Mandar Joshi, Danqi Chen, Omer Levy, Mike Lewis, Luke Zettlemoyer, and Veselin Stoyanov. 2019.
\newblock Roberta: A robustly optimized bert pretraining approach.
\newblock \emph{arXiv preprint arXiv:1907.11692}.

\bibitem[{Lucas et~al.(2023)Lucas, Uchendu, Yamashita, Lee, Rohatgi, and Lee}]{lucas2023fighting}
Jason Lucas, Adaku Uchendu, Michiharu Yamashita, Jooyoung Lee, Shaurya Rohatgi, and Dongwon Lee. 2023.
\newblock Fighting fire with fire: The dual role of llms in crafting and detecting elusive disinformation.
\newblock In \emph{Proceedings of the 2023 Conference on Empirical Methods in Natural Language Processing}, pages 14279--14305.

\bibitem[{Macko et~al.(2024)Macko, Moro, Uchendu, Srba, Lucas, Yamashita, Tripto, Lee, Simko, and Bielikov{\'a}}]{macko2024authorship}
Dominik Macko, Robert Moro, Adaku Uchendu, Ivan Srba, Jason Lucas, Michiharu Yamashita, Nafis~Irtiza Tripto, Dongwon Lee, Jakub Simko, and M{\'a}ria Bielikov{\'a}. 2024.
\newblock Authorship obfuscation in multilingual machine-generated text detection.
\newblock In \emph{Findings of the Association for Computational Linguistics: EMNLP 2024}, pages 6348--6368.

\bibitem[{Mahmood et~al.(2019)Mahmood, Ahmad, Shafiq, Srinivasan, and Zaffar}]{mahmood2019girl}
Asad Mahmood, Faizan Ahmad, Zubair Shafiq, Padmini Srinivasan, and Fareed Zaffar. 2019.
\newblock A girl has no name: Automated authorship obfuscation using mutant-x.
\newblock \emph{Proceedings on Privacy Enhancing Technologies}.

\bibitem[{Mir et~al.(2019)Mir, Felbo, Obradovich, and Rahwan}]{mir2019evaluating}
Remi Mir, Bjarke Felbo, Nick Obradovich, and Iyad Rahwan. 2019.
\newblock Evaluating style transfer for text.
\newblock In \emph{Proceedings of the 2019 Conference of the North American Chapter of the Association for Computational Linguistics: Human Language Technologies, Volume 1 (Long and Short Papers)}, pages 495--504.

\bibitem[{Nagrani et~al.(2017)Nagrani, Chung, and Zisserman}]{nagrani2017vox}
Arsha Nagrani, Joon~Son Chung, and Andrew Zisserman. 2017.
\newblock \href {https://arxiv.org/abs/1706.08612} {Voxceleb: a large-scale speaker identification dataset}.
\newblock \emph{CoRR}, abs/1706.08612.

\bibitem[{Nguyen et~al.(2023)Nguyen, Alperin, Dagli, Vandam, and Singer}]{nguyen2023improving}
Trang Nguyen, Kenneth Alperin, Charlie Dagli, Courtland Vandam, and Elliot Singer. 2023.
\newblock Improving long-text authorship verification via model selection and data tuning.
\newblock In \emph{Proceedings of the 7th Joint SIGHUM Workshop on Computational Linguistics for Cultural Heritage, Social Sciences, Humanities and Literature}, pages 28--37.

\bibitem[{Oak(2022)}]{oak2022poster}
Rajvardhan Oak. 2022.
\newblock Poster--towards authorship obfuscation with language models.
\newblock In \emph{Proceedings of the 2022 ACM SIGSAC Conference on Computer and Communications Security}, pages 3435--3437.

\bibitem[{Potha and Stamatatos(2014)}]{potha2014profile}
Nektaria Potha and Efstathios Stamatatos. 2014.
\newblock A profile-based method for authorship verification.
\newblock In \emph{Hellenic Conference on Artificial Intelligence}, pages 313--326. Springer.

\bibitem[{Potthast et~al.(2016)Potthast, Hagen, and Stein}]{potthast2016author}
Martin Potthast, Matthias Hagen, and Benno Stein. 2016.
\newblock Author obfuscation: Attacking the state of the art in authorship verification.
\newblock \emph{CLEF (Working Notes)}, pages 716--749.

\bibitem[{Qi et~al.(2021)Qi, Chen, Zhang, Li, Liu, and Sun}]{qi2021mind}
Fanchao Qi, Yangyi Chen, Xurui Zhang, Mukai Li, Zhiyuan Liu, and Maosong Sun. 2021.
\newblock Mind the style of text! adversarial and backdoor attacks based on text style transfer.
\newblock In \emph{Proceedings of the 2021 Conference on Empirical Methods in Natural Language Processing}, pages 4569--4580.

\bibitem[{Ramnath et~al.(2024)Ramnath, Pandey, Boschee, and Ren}]{ramnath2024cave}
Sahana Ramnath, Kartik Pandey, Elizabeth Boschee, and Xiang Ren. 2024.
\newblock Cave: Controllable authorship verification explanations.
\newblock \emph{arXiv preprint arXiv:2406.16672}.

\bibitem[{Ren et~al.(2019)Ren, Deng, He, and Che}]{ren2019generating}
Shuhuai Ren, Yihe Deng, Kun He, and Wanxiang Che. 2019.
\newblock Generating natural language adversarial examples through probability weighted word saliency.
\newblock In \emph{Proceedings of the 57th annual meeting of the association for computational linguistics}, pages 1085--1097.

\bibitem[{Seidman(2013)}]{seidman2013authorship}
Shachar Seidman. 2013.
\newblock Authorship verification using the impostors method.
\newblock In \emph{CLEF 2013 Evaluation labs and workshop--Working notes papers}, pages 23--26.

\bibitem[{Shetty et~al.(2018)Shetty, Schiele, and Fritz}]{shetty2018a4nt}
Rakshith Shetty, Bernt Schiele, and Mario Fritz. 2018.
\newblock $\{$A4NT$\}$: Author attribute anonymity by adversarial training of neural machine translation.
\newblock In \emph{27th USENIX Security Symposium (USENIX Security 18)}, pages 1633--1650.

\bibitem[{Singer et~al.(2023)Singer, Borgstr{\"o}m, Alperin, Nguyen, Dagli, Dale, and Ross}]{singer2023design}
Elliot Singer, Bengt~J Borgstr{\"o}m, Kenneth Alperin, Trang Nguyen, Cagri Dagli, Melissa Dale, and Arun Ross. 2023.
\newblock On the design of the mitll trimodal dataset for identity verification.
\newblock In \emph{2023 11th International Workshop on Biometrics and Forensics (IWBF)}, pages 1--6. IEEE.

\bibitem[{Stamatatos(2009)}]{stamatatos2009survey}
Efstathios Stamatatos. 2009.
\newblock A survey of modern authorship attribution methods.
\newblock \emph{Journal of the American Society for information Science and Technology}, 60(3):538--556.

\bibitem[{Stamatatos(2016)}]{stamatatos2016authorship}
Efstathios Stamatatos. 2016.
\newblock Authorship verification: A review of recent advances.
\newblock \emph{Res. Comput. Sci.}, 123(1):9--25.

\bibitem[{Steinfeld(2022)}]{steinfeld2022disinformation}
Nili Steinfeld. 2022.
\newblock The disinformation warfare: how users use every means possible in the political battlefield on social media.
\newblock \emph{Online Information Review}, 46(7):1313--1334.

\bibitem[{Tripto et~al.(2023)Tripto, Uchendu, Le, Setzu, Giannotti, and Lee}]{tripto2023hansen}
Nafis~Irtiza Tripto, Adaku Uchendu, Thai Le, Mattia Setzu, Fosca Giannotti, and Dongwon Lee. 2023.
\newblock Hansen: Human and ai spoken text benchmark for authorship analysis.
\newblock In \emph{The 2023 Conference on Empirical Methods in Natural Language Processing}.

\bibitem[{Tyo et~al.(2022)Tyo, Dhingra, and Lipton}]{tyo2022state}
Jacob Tyo, Bhuwan Dhingra, and Zachary~C Lipton. 2022.
\newblock On the state of the art in authorship attribution and authorship verification.
\newblock \emph{arXiv preprint arXiv:2209.06869}.

\bibitem[{Uchendu et~al.(2023)Uchendu, Le, and Lee}]{uchendu2023attribution}
Adaku Uchendu, Thai Le, and Dongwon Lee. 2023.
\newblock Attribution and obfuscation of neural text authorship: A data mining perspective.
\newblock \emph{ACM SIGKDD Explorations Newsletter}, 25(1):1--18.

\bibitem[{Valdez-Valenzuela and G{\'o}mez-Adorno(2024)}]{valdez2024team}
Andric Valdez-Valenzuela and Helena G{\'o}mez-Adorno. 2024.
\newblock Team iimasnlp at pan: leveraging graph neural networks and large language models for generative ai authorship verification.
\newblock \emph{Working Notes of CLEF}.

\bibitem[{Wang et~al.(2024)Wang, Shi, Bai, and Hsieh}]{wang2024defending}
Yihan Wang, Zhouxing Shi, Andrew Bai, and Cho-Jui Hsieh. 2024.
\newblock Defending llms against jailbreaking attacks via backtranslation.
\newblock In \emph{Findings of the Association for Computational Linguistics ACL 2024}, pages 16031--16046.

\bibitem[{Weerasinghe et~al.(2021)Weerasinghe, Singh, and Greenstadt}]{weerasinghe2021feature}
Janith Weerasinghe, Rhia Singh, and Rachel Greenstadt. 2021.
\newblock Feature vector difference based authorship verification for open-world settings.
\newblock In \emph{CLEF (Working Notes)}, pages 2201--2207.

\bibitem[{Wiegmann et~al.(2019)Wiegmann, Stein, and Potthast}]{wiegmann2019celebrity}
Matti Wiegmann, Benno Stein, and Martin Potthast. 2019.
\newblock \href {https://doi.org/10.18653/v1/P19-1249} {Celebrity profiling}.
\newblock In \emph{Proceedings of the 57th Annual Meeting of the Association for Computational Linguistics}, pages 2611--2618, Florence, Italy. Association for Computational Linguistics.

\bibitem[{Xing et~al.(2024)Xing, Venkatraman, Le, and Lee}]{xing2024alison}
Eric Xing, Saranya Venkatraman, Thai Le, and Dongwon Lee. 2024.
\newblock Alison: Fast and effective stylometric authorship obfuscation.
\newblock In \emph{Proceedings of the AAAI Conference on Artificial Intelligence}, volume~38, pages 19315--19322.

\bibitem[{Zhang et~al.(2020)Zhang, Zhao, Saleh, and Liu}]{zhang2020pegasus}
Jingqing Zhang, Yao Zhao, Mohammad Saleh, and Peter Liu. 2020.
\newblock Pegasus: Pre-training with extracted gap-sentences for abstractive summarization.
\newblock In \emph{International conference on machine learning}, pages 11328--11339. PMLR.

\end{thebibliography}


\appendix

\begin{figure*}[!htb]
    \centering
    \includegraphics[width=1\linewidth]{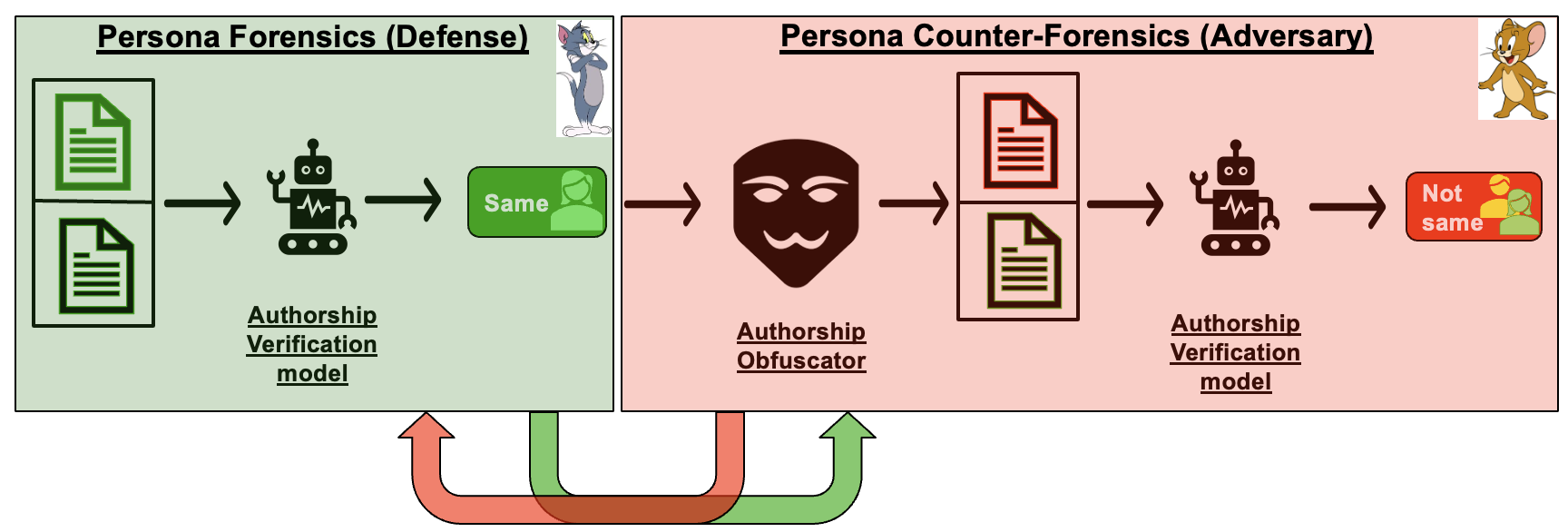}
    \caption{Methodological framework for Authorship Obfuscation of the Authorship Verification model}
    \label{fig:framework}
\end{figure*}

\section{Methodology}

\subsection{Authorship Obfuscation}
See Figure \ref{fig:framework} for a flowchart that illustrates the process of obfuscating the writing style of a pair\_1 
from two pairs of a document that have been accurately verified as written by the same author.  
See Table \ref{tab:obf_prompts} for the prompts we used with Mistral to construct our obfuscation attacks.

\begin{figure*}[!htb]
    \centering
    \fbox{
    \includegraphics[width=1\linewidth]{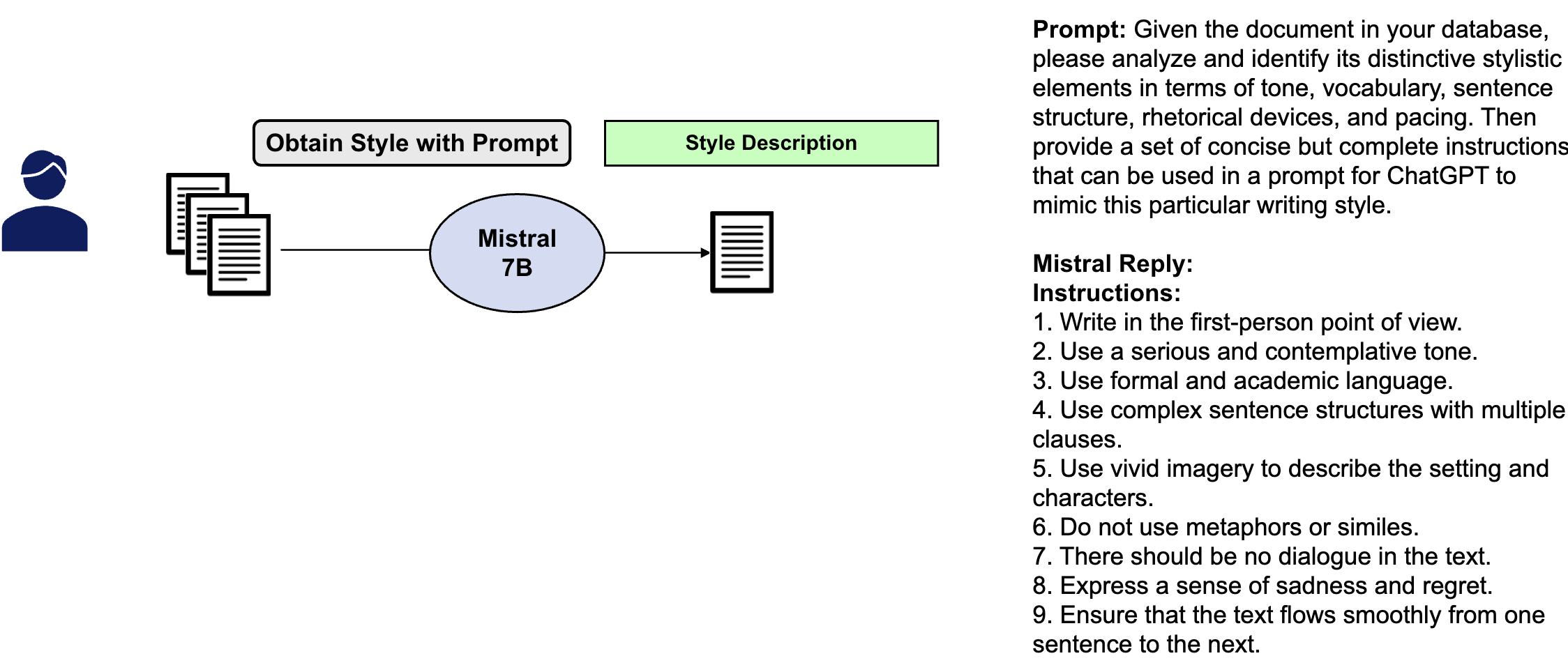}}
    \fbox{\includegraphics[width=1\linewidth]{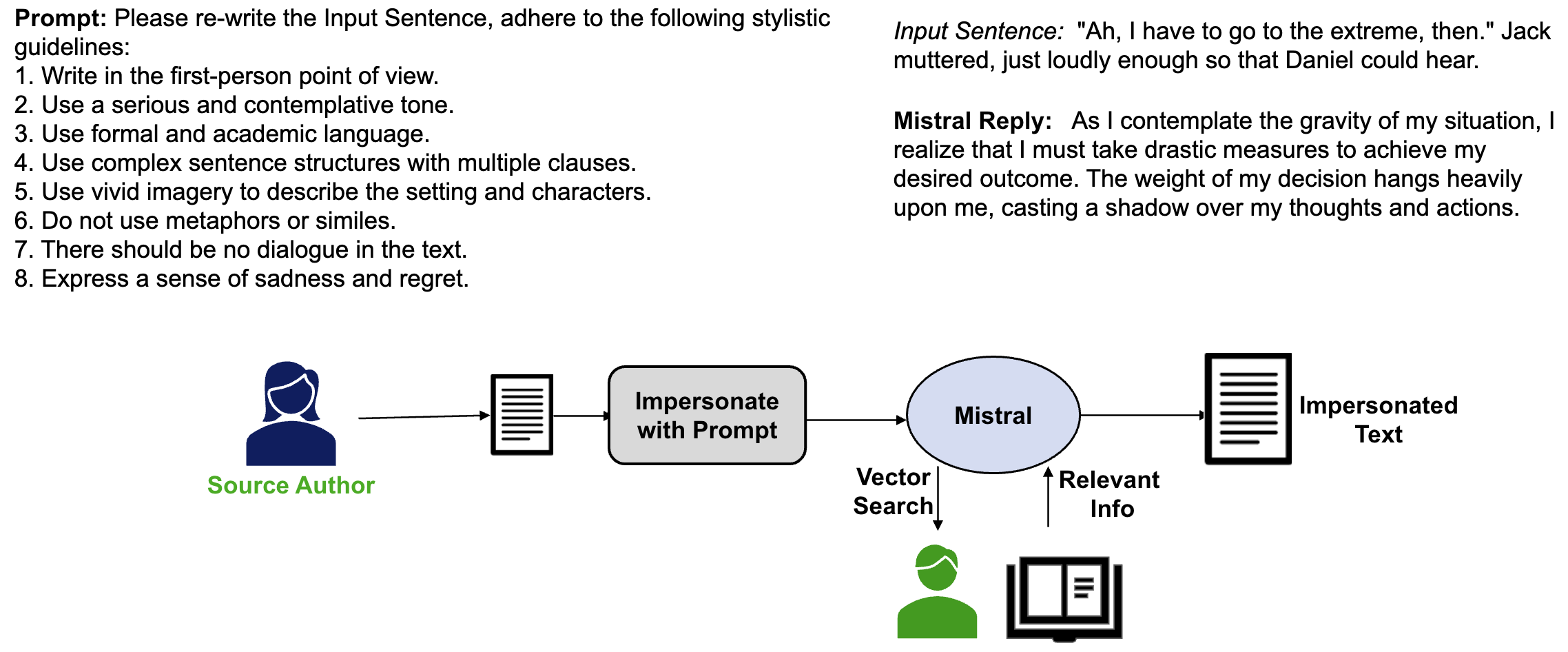}}
    \caption{RAG Pipeline with Mistral}
    \label{fig:rag_mistral}
\end{figure*}

\subsection{Authorship Impersonation}
We perform impersonation attacks, with the following methods: 

\begin{itemize}
    \item \textbf{Mistral and RAG}:
    Our multi-step RAG pipeline is structured as follows:
    In the first part of our multi-step RAG pipeline we collect a dataset containing the writings of the target author, each piece of text is converted into vector embeddings using a MPNET embedding model. These embeddings represent the semantic and stylistic properties of the author's writing in a high-dimensional space. We then use the Mistral model to query the target author’s embedding to retrieve the style descriptions of the target author. We use chain of thought (COT) prompting to retrieve the stylistic description of the target author. COT helps in breaking down the prompt into step-by-step instructions allowing the better search and retrieval. The output from the CoT-augmented retrieval provides a structured summary of the target author's style, such as sentence structure, vocabulary, tone, and rhythm. 
    In the next part of our multi-step RAG pipeline, we use the style summary generated from the first step of our pipeline, to guide the Mistral-7B v0.1 model to rewrite the source author’s content while maintaining its original meaning. This implies that while the content of writing of the source author remains the same, the style, tone and structure reflect that of the target author. See Figure \ref{fig:rag_mistral} for an illustration of the RAG pipeline infused with Mistral for authorship impersonation. 

    \item \textbf{STRAP}: We perform authorship impersonation using the STRAP (Style Transfer via Paraphrasing) framework introduced by \cite{krishna2020reformulating}. The pipeline involves three key phases: paraphrasing, fine-tuning, and style imputation. 
    \begin{itemize}
    
    \item \textit{Paraphrasing with STRAP}:
    The first step in our pipeline employs the STRAP framework to generate paraphrased versions of both the source and target author documents. STRAP reformulates unsupervised style transfer as a paraphrase generation task, where the style of a given sentence is modified without significantly altering its meaning. This process involves feeding the original sentences through a diverse paraphrase model to create pseudo-parallel datasets. 

    \item \textit{Fine-tuning GPT-2 for Style Transfer}:
    Once the paraphrased documents are generated, we use them alongside their original counterparts to fine-tune a GPT-2 model. This fine-tuning process allows the model to learn the relationship between neutral, paraphrased text and its corresponding author-specific style. The GPT-2 model is fine-tuned to learn stylistic features that are characteristic of the target author.

    \item \textit{Style Imputation on Source Text}:
    After fine-tuning, the GPT-2 model is applied to the paraphrased source documents. This step, referred to as style imputation or inferencing, involves passing the paraphrased source author texts from Step 1 to the fine-tuned GPT-2 model allowing it to adopt the stylistic features of the target author. This allows for the source author text to adapt the style of the target author without changing its semantics.
    
    \end{itemize}
    
\end{itemize}

\begin{table*}[!htb]
\footnotesize
    \centering
    \begin{tabular}{p{0.20\linewidth} p{0.70\linewidth}}
\hline
\textbf{Attack Technique} & \textbf{Prompt} \\
\hline
$Mistral\_{vanilla}$ & 
\begin{minipage}[t]{10cm}
Can you paraphrase the text: \{target\_article\}?
\\Please only provide the paraphrased text and provide no other explanation for your answer. \\Only respond with the text that has been paraphrased. 
\end{minipage}
\\
\hline

$Mistral\_{zeroshot}$ &
\begin{minipage}[t]{10cm}
You are an expert hacker and linguist. The purpose of paraphrasing writing style of a text while maintaining its semantics is to make language models more robust to adversarial attacks. Your task is to generate a new sentence which must satisfy the following conditions:\\
1. The new sentence should paraphrase 30\% of the original sentence \\
2. Keeping the semantic meaning of the new sentence unchanged; \\
\\
You can finish the task by modifying the sentence using the following guidance: \\
1. Paraphrase at most 30\% of the original sentence \\
2. Only output the new sentence without anything else.
\end{minipage} \\
\hline

$Mistral\_{stepback}$ &
\begin{minipage}[t]{10cm}
You are an expert writer and linguist. \\
Question: What is the purpose of obfuscating the writing style of a text while maintaining its
semantics to fool a classification model? \\
Answer:  The purpose of obfuscating the writing style of a text while maintaining its 
semantics is to make it more difficult for a classification model to accurately predict the category or label of the text. This is often done in the context of natural language processing (NLP) and machine learning, where the goal is to train a model to accurately classify texts based on their content. Obfuscating the writing style of a text can make it more difficult for the model to identify patterns or features that are specific to a particular category or label. This can be done by using techniques such as replacing common words with synonyms, using jargon or technical terms, or using complex sentence structures. By making it more difficult for the model to accurately classify the text, it can improve the model's overall performance and reduce the risk of overfitting, which occurs when the model becomes too specialized to the training data and performs poorly on new, unseen data. Overall, obfuscating the writing style of a text can help to improve the accuracy and robustness of a classification 
model, making it better able to handle a wider range of inputs and accurately predict their categories or labels.
 \\
 \\
</s>
Question: Can you Obfuscate writing style of the text while keeping semantics as similar as possible to fool a classification model so that it can convert a true trial two texts written by same author identified correctly, into a false trial two texts written by same author as incorrectly
\end{minipage} \\
\hline

$Mistral\_{AP}$ &
\begin{minipage}[t]{10cm}
 \{lexical\_div\} definition. You are an expert hacker and linguist. Your task is \\
 1. Increase lexical diversity by 60\% by utilizing the following stylistic elements used to write -voice, tone, diction, sentence structure, metaphors \& similes, pacing, imagery, dialogue, age-related features, gender-related features, educational background, psychological traits, cultural \& geographic influences, and social \& occupational factors \\
 2. Keep semantics the same and in the modern era \\
 3. Paraphrase only at most 30\% of the text \\
Make sure you only output the new diverse sentence and nothing else, no explanation.
Using the instructions paraphrase this text: \{target\_article\} 
\end{minipage} \\

\hline
\end{tabular}
    \caption{Authorship Obfuscation prompts}
    \label{tab:obf_prompts}
\end{table*}



\end{document}